%% file: main.tex
\documentclass[11pt]{article}
\usepackage[inline]{enumitem}
\usepackage[utf8]{inputenc}
\usepackage[preprint]{automl} % preprint

\makeatletter
\renewcommand{\c@nferencestring}{
    \raisebox{2ex}[0pt][0pt]{
        \parbox[t]{0.5\textwidth}{
            \raggedright\scriptsize An earlier version of this paper was published at AutoML 2025. This version adds missing standard deviations in Table~1.
        }
    }
}
\makeatother

\usepackage{siunitx}
\input{biblatex_macro}

\usepackage{tabularray}
\usepackage{adjustbox}
\usepackage{xspace}
\usepackage{stackengine}
\usepackage[np]{numprint}
\usepackage{alltt}
\usepackage{graphicx}
\usepackage{listings}
\usepackage{mathpartir}
\usepackage{marginalia}
\usepackage{algorithm}
\usepackage{algorithmic}
\usepackage{wrapfig}
\usepackage{flafter}

\usepackage{tcolorbox}
\tcbuselibrary{listings,breakable}

\newcounter{rqcounter}
\renewcommand{\therqcounter}{RQ~\arabic{rqcounter}}

\newtcolorbox{rqbox}[1][]{%
  colback=gray!20,
  colframe=black,
  fonttitle=\bfseries,
  fontupper=\small,
  top=0pt,
  bottom=0pt,
  left=0pt,
  right=0pt,
  #1
}

\newtcolorbox{findingsbox}[1][]{%
  colback=gray!20,
  colframe=black,
  fonttitle=\bfseries,
  fontupper=\small,
  top=0pt,
  bottom=0pt,
  left=0pt,
  right=0pt,
  #1
}

\newcommand{\xyz}[1]{
\begin{rqbox}
\textbf{~\refstepcounter{rqcounter}\therqcounter}: #1
\end{rqbox}}

\DeclareMathOperator*{\argmin}{argmin}

\usepackage{hyperref}

\usepackage[capitalize,nameinlink]{cleveref}
\usepackage{aliascnt}

\newaliascnt{appfigure}{figure}
\crefname{appfigure}{Appendix Figure}{Appendix Figures}
\Crefname{appfigure}{Appendix Figure}{Appendix Figures}

\hypersetup{%
  pdfauthor={}, % will be reset to "Anonymous" unless the "final" package option is given
  pdftitle={},
  pdfsubject={},
  pdfkeywords={},
  colorlinks=true,
  citecolor=blue,
  linkcolor=blue,
  urlcolor=blue
}

\input{macros}

\definecolor{keyword}{HTML}{37AC4A}
\definecolor{jinja2}{HTML}{0070C0}
\definecolor{comment}{HTML}{7F7F7F}
\definecolor{generated}{HTML}{37AC4A}
\definecolor{toolout}{HTML}{A51DFF}

\lstdefinelanguage{pdl}{
  xleftmargin=4mm,
  numbers=left,
  numbersep=2mm,
  numberstyle=\tiny\color{gray},
  basicstyle=\fontsize{8}{9}\ttfamily,
  basewidth=0.5em,
  alsoletter={:},
  morekeywords={array:,as:call:,code:,content:,contribute:,data:,def:,defs:,description:,else:,for:,function:,if:,include:,input:,join:,lang:,lastOf:,message:,model:,multiline:,num_iterations:,object:,parameters:,parser:,pdl_context:,read:,repeat:,return:,role:,spec:,text:,then:,until:,with:,prompt_pattern:,num_demonstrations:,system_prompt:,question:,thought:,action:,observation:,demonstrations:,call:,args:,task:,tools:,trajectories:,prompt_pattern:,reasoning:,answer:,examples:,instruction:},
  keywordstyle=\color{keyword}\bf,
  morestring=[s]{$\{}{\}},
  stringstyle=\color{jinja2},
  emphstyle=\color{jinja2},
  showstringspaces=false,
  morecomment=[l]{\#},
  commentstyle=\color{comment}\it,
}

\lstdefinelanguage{myyaml}{
  xleftmargin=4mm,
  numbers=left,
  numbersep=2mm,
  numberstyle=\tiny\color{gray},
  basicstyle=\fontsize{8}{9}\ttfamily,
  basewidth=0.5em,
  alsoletter={:},
  morekeywords={initial_test_set_size:,max_test_set_size:,num_candidates:,parallelism:,variables:,prompt_pattern:,num_demonstrations:,system_prompt:,demonstrations:},
  keywordstyle=\color{keyword}\bf,
  morestring=[s]{\$\{}{\}},
  stringstyle=\color{jinja2},
  emphstyle=\color{jinja2},
  showstringspaces=false,
  morecomment=[l]{\#},
  commentstyle=\color{comment}\it,
  moredelim=[is][commentstyle]{||}{££}, % invisible custom delimiters
  sensitive=false,
  morestring=[b]',
  morestring=[b]"
}

\begin{document}
\title{AutoPDL: Automatic Prompt Optimization for LLM Agents}
\author[1]{\nameemail{Claudio Spiess}{cvspiess@ucdavis.edu}}
\author[2]{\nameemail{Mandana Vaziri}{mvaziri@us.ibm.com}}
\author[2]{\nameemail{Louis Mandel}{lmandel@us.ibm.com}}
\author[2]{\nameemail{Martin Hirzel}{hirzel@us.ibm.com}}

\affil[1]{UC Davis}
\affil[2]{IBM Research}

\maketitle
\begin{abstract}
\input{abstract.tex}
\end{abstract}

\section{Introduction}
\input{sections/01_intro}

\section{Background}
\input{sections/02_background}

\section{AutoPDL Approach}
\input{sections/03_approach}

\section{Methodology}
\input{sections/04_methodology}

\section{Results}\label{sec:results}
\input{sections/05_results}

\section{Related Work}
\input{sections/06_related}

\section{Conclusion}
\input{sections/07_conclusion}

\clearpage
\printbibliography % BibLaTeX
% \bibliography{main.bib}

\clearpage

\input{sections/08_appendix}

\end{document}

%% file: biblatex_macro.tex
\PassOptionsToPackage{
        natbib=false,
        style=authoryear-comp,
        hyperref=true,
        backend=biber,
        maxbibnames=99,
        giveninits=true,
        uniquename=init,
        maxcitenames=1,
        parentracker=true,
        url=false,
        doi=false,
        isbn=false,
        eprint=false,
        backref=true,
            }{biblatex}
\usepackage{biblatex}
\DeclareNameAlias{sortname}{family-given}

\renewbibmacro{in:}{%
  \ifentrytype{article}{}{%
  \printtext{\bibstring{in}\intitlepunct}}}

\AtEveryBibitem{%
  \clearfield{month}{}%
  \clearlist{language}{}%
  }

\let\cite\parencite

\renewbibmacro*{volume+number+eid}{%
  \setunit*{\addcomma\space}% NEW
  \printfield{volume}%
  \printfield{number}%
  \printfield{eid}}
  \DeclareFieldFormat[article]{number}{(#1)}% number of a journal

\makeatletter
\renewbibmacro*{cite}{% Based on cite bib macro from authoryear-comp.cbx
  \iffieldundef{shorthand}
    {\ifthenelse{\ifnameundef{labelname}\OR\iffieldundef{labelyear}}
       {\printtext[bibhyperref]{% Include labelname in hyperlink
          \DeclareFieldAlias{bibhyperref}{default}% Prevent nested hyperlinks
          \usebibmacro{cite:label}%
          \setunit{\addspace}%
          \usebibmacro{cite:labeldate+extradate}}%
          \usebibmacro{cite:reinit}}
       {\iffieldequals{namehash}{\cbx@lasthash}
          {\ifthenelse{\iffieldequals{labelyear}{\cbx@lastyear}\AND
                       \(\value{multicitecount}=0\OR\iffieldundef{postnote}\)}
             {\setunit{\addcomma}%
              \usebibmacro{cite:extrayear}}
             {\setunit{\compcitedelim}%
              \usebibmacro{cite:labeldate+extradate}%
              \savefield{labelyear}{\cbx@lastyear}}}
          {\printtext[bibhyperref]{% Include labelname in hyperlink
             \DeclareFieldAlias{bibhyperref}{default}% Prevent nested hyperlinks
             \printnames{labelname}%
             \setunit{\nameyeardelim}%
             \usebibmacro{cite:labeldate+extradate}}%
             \savefield{namehash}{\cbx@lasthash}%
             \savefield{labelyear}{\cbx@lastyear}}}}
    {\usebibmacro{cite:shorthand}%
     \usebibmacro{cite:reinit}}%
  \setunit{\multicitedelim}}

\renewbibmacro*{textcite}{% Based on textcite bib macro from authoryear-comp.cbx
  \iffieldequals{namehash}{\cbx@lasthash}
    {\iffieldundef{shorthand}
       {\ifthenelse{\iffieldequals{labelyear}{\cbx@lastyear}\AND
                    \(\value{multicitecount}=0\OR\iffieldundef{postnote}\)}
          {\setunit{\addcomma}%
           \usebibmacro{cite:extrayear}}
          {\setunit{\compcitedelim}%
           \usebibmacro{cite:labeldate+extradate}%
           \savefield{labelyear}{\cbx@lastyear}}}
       {\setunit{\compcitedelim}%
        \usebibmacro{cite:shorthand}%
        \global\undef\cbx@lastyear}}
    {\ifnameundef{labelname}
       {\printtext[bibhyperref]{% Include labelname in hyperlink
          \DeclareFieldAlias{bibhyperref}{default}% Prevent nested hyperlinks
          \iffieldundef{shorthand}
            {\usebibmacro{cite:label}%
             \setunit{%
               \global\booltrue{cbx:parens}%
               \addspace\bibopenparen}%
             \ifnumequal{\value{citecount}}{1}
               {\usebibmacro{prenote}}
               {}%
             \usebibmacro{cite:labeldate+extradate}}
            {\usebibmacro{cite:shorthand}}%
          \ifthenelse{\iffieldundef{postnote}\AND
                      \(\value{multicitetotal}=0\AND\value{citetotal}=1\)}
            {\bibcloseparen% Include closing parenthesis in hyperlink
             \global\boolfalse{cbx:parens}}
            {}}}
       {\printtext[bibhyperref]{% Include labelname in hyperlink
          \DeclareFieldAlias{bibhyperref}{default}% Prevent nested hyperlinks
          \printnames{labelname}%
          \setunit{%
            \global\booltrue{cbx:parens}%
            \addspace\bibopenparen}%
          \ifnumequal{\value{citecount}}{1}
            {\usebibmacro{prenote}}
            {}%
          \iffieldundef{shorthand}
            {\iffieldundef{labelyear}
               {\usebibmacro{cite:label}}
               {\usebibmacro{cite:labeldate+extradate}}%
             \savefield{labelyear}{\cbx@lastyear}}
            {\usebibmacro{cite:shorthand}%
             \global\undef\cbx@lastyear}%
          \ifthenelse{\iffieldundef{postnote}\AND
                      \(\value{multicitetotal}=0\AND\value{citetotal}=1\)}
            {\bibcloseparen% Include closing parenthesis in hyperlink
             \global\boolfalse{cbx:parens}}
            {}}%
          \savefield{namehash}{\cbx@lasthash}}}%
  \setunit{%
    \ifbool{cbx:parens}
      {\bibcloseparen\global\boolfalse{cbx:parens}}
      {}%
    \multicitedelim}}

\makeatother

%% file: macros.tex
\newcommand{\ie}{\emph{i.e.},\xspace}
\newcommand{\eg}{\emph{e.g.},\xspace}
\newcommand{\vs}{\emph{vs.}\xspace}

\newcommand{\sref}[1]{\S~\ref{#1}}

%% file: abstract.tex
The performance of large language models (LLMs) depends on how they are prompted, with choices spanning both the high-level prompting pattern (e.g., Zero-Shot, CoT, ReAct, ReWOO) and the specific prompt content (instructions and few-shot demonstrations).
Manually tuning this combination is tedious, error-prone, and specific to a given LLM and task.
Therefore, this paper proposes AutoPDL, an automated approach to discovering good LLM agent configurations.
Our approach frames this as a structured AutoML problem over a combinatorial space of agentic and non-agentic prompting patterns and demonstrations, using successive halving to efficiently navigate this space.
We introduce a library implementing common prompting patterns using the PDL prompt programming language.
\mbox{AutoPDL} solutions are human-readable, editable, and executable PDL programs that use this library.
This approach also enables source-to-source optimization, allowing human-in-the-loop refinement and reuse.
Evaluations across three tasks and seven LLMs (ranging from 3B to 70B parameters) show consistent accuracy gains ($9.21\pm15.46$ percentage points), up to 67.5pp, and reveal that selected prompting strategies vary across models and tasks.

%% file: sections/01_intro.tex
Large language models (LLMs) and LLM-based agents excel at a variety of tasks, including question answering, math
word problems, and programming.
The performance of an LLM depends heavily on how it is prompted, and
there are a variety of popular prompting patterns.
These include zero-shot or few-shot~\cite{brown_et_al_2020} prompting % Claudio: succinct fewshot intro from https://www.promptingguide.ai/techniques/fewshot = Few-shot prompting can be used as a technique to enable in-context learning, where we provide demonstrations in the prompt to steer the model towards better performance. The demonstrations serve as conditioning for subsequent queries where we would like the model to generate a response.
with chain-of-thought~(CoT)~\cite{weiChainThoughtPromptingElicits2023}, zero-shot CoT~\cite{kojimaLargeLanguageModels2022}, 
as well as agentic patterns such as ReAct~\cite{yao_et_al_2023} or
ReWOO~\cite{xuReWOODecouplingReasoning2023}.
However, given a dataset $D_\textrm{test}$ with a loss function
$\mathcal{L}$, \eg error rate, it is not clear which pattern will do best.
Furthermore, besides the pattern~$A$, performance also depends on the
prompt~$p$, including few-shot samples and instructions.
The problem is thus to find a combination $A_p^*$ of a pattern along
with an optimized prompt that minimizes~$\mathcal{L}$.
This is usually done via manual prompt engineering, but that is
tedious and has to be repeated if a new LLM comes along.
Therefore, this paper explores how to find $A_p^*$ using automated
machine learning~(AutoML).
And for users to trust the result or to tweak it further,
$A_p^*$ itself should be easy to read and edit.

Agent frameworks, such as CrewAI~\cite{moura_2023} or
AutoGen~\cite{wu_et_al_2023}, contain pre-built agent patterns with
prompts optimized for proprietary frontier models and common tasks.
Unfortunately, their prompts are deeply buried~\cite{schluntz_zhang_2024},
making them hard to modify and adapt to non-frontier models or novel tasks.
Moreover, the prompting pattern is fixed to a variation of ReAct, limiting flexibility in customizing the prompt structure.
Prompt optimizers, such as DSPy~\cite{khattab_et_al_2024}, optimize
few-shot samples for in-context learning (ICL) and/or instructions
in the prompt~$p$.
Unfortunately, they do not automatically select the prompting
pattern~$A$ and do not return human-readable code.

We formulate the problem of finding a good pattern and corresponding
prompt by defining and then exploring a combined search space.
We were inspired by the AutoML literature on combined search spaces of
machine-learning algorithms and their hyperparameters~\cite{thornton_et_al_2013},
except that (i)~instead of discrete or continuous hyperparameters, we
explore textual ICL samples, instructions, and prompting patterns;
(ii)~instead of classification or regression tasks, we tackle generative tasks;
and (iii)~instead of model training or fine-tuning, we focus on in-context learning.
We assume a dataset with a validation set $D_\mathrm{valid}$, test set $D_\mathrm{test}$,
as well as an example bank $D_\mathrm{train}$ for few-shot samples. $D_\mathrm{valid}$ is used during the optimization process to evaluate the performance of a configuration, whereas $D_\mathrm{test}$ is used once upon completion, to evaluate the performance of the best performing configuration.
As usual, to avoid over-fitting, we assume these are disjoint from each other.
The problem statement is to find
$\displaystyle A_p^* = \argmin_{A_p\in\mathcal{A}_\mathcal{P}}\mathcal{L}(A_p,D_\textrm{valid})$, where:
\begin{itemize}
  \item $A\in\mathcal{A} = \{\textrm{Zero-Shot, CoT, ReWOO, ReAct}\}$
  is the prompting pattern, and
  \item $p=\langle n,d_\textrm{train},\textrm{instr}\rangle\in\mathcal{P}$
  is the prompt, comprising
  a number $n\le|D_\mathrm{train}|$ of few-shot samples,
  the actual few-shot samples $d_\mathrm{train}\in (D_\mathrm{train})^n$,
  and an instruction $\textrm{instr}\in\mathcal{I}$. A concrete example of $\mathcal{P}$ is provided in~\autoref{fig:pdlConcrete} and~\autoref{fig:pdlConcreteTxt}.
  %TODO: Instr is unclear. In the end it was only used in ReAct. Remove? -> I would keep it for completeness
\end{itemize}

\noindent
To avoid getting stuck in local minima while saving compute and finding
a solution with a low loss, we explore the search
space~$\mathcal{A}_\mathcal{P}$ using successive halving~\cite{jamieson_talwalkar_2016}.
To make the initial search space $\mathcal{A}_\mathcal{P}$ user-interpretable,
and the final solution $A_p^*$ both human readable and executable, we express them
in a YAML-based prompting language, PDL~\cite{vaziriPDLDeclarativePrompt2024}. PDL's structured format makes it easy to modify both the initial search space and the optimized program, and ensures the final solution remains directly executable.
We introduce a library for PDL that implements each of the
common prompting patterns in~$\mathcal{A}$.
The initial search space $\mathcal{A}_\mathcal{P}$ is a YAML
file with various choice points for AutoML to decide.
And the solution $A_p^*$ is a custom-tailored PDL program optimized
for the given task, as given by the dataset and loss function.
The developer can read or even tweak either or both as desired.

We evaluate our optimizer on three tasks (question answering, math,
and programming), using seven LLMs sized between 3B and 70B parameters.
We find that the optimizer often gives accuracy boosts in the 6--30\%
range, in some cases higher.
Given the same task, different patterns $A\in\mathcal{A}$ do best for
different models. 
%(There were a few rare cases where the zero-shot baseline wins.)
Conversely, given the same model, different patterns do best for
different tasks.
Besides this variability in the chosen pattern $A$, our experiments
also revealed variability in the optimized
prompts~$p=\langle n,d_\textrm{train},\textrm{instr}\rangle$.
We also found that when training data for a task is missing, data from
a related but different dataset can help.
Also, while most of our experiments use moderately-sized open models, we
also show our optimized solutions can benefit frontier models.

This paper makes three primary contributions:
\begin{enumerate}
    \item Jointly searching pattern and prompt: prior work in prompt optimization has not investigated searching \emph{joint} search spaces, including agentic patterns.
    \item No one size fits all: we find that different models sometimes have differing optimal prompt patterns for the \emph{same} benchmark, suggesting that there is not one single optimal prompt pattern.
    \item Source-to-source optimization: we propose the first source-to-source optimizer for LLM prompt programs, where both the initial search space and the final solution are prompt programs in the same language, making the final solution both human-readable and executable.
\end{enumerate}

% TODO: add a pointer to the open-source code, while also mentioning the artifact.
Overall, this paper shows how to apply AutoML to automatically
discover agentic or non-agentic LLM prompts and patterns optimized for a given task. We make our AutoPDL implementation available at~\url{https://github.com/IBM/prompt-declaration-language}, and release the \href{https://openreview.net/attachment?id=CAeISyE3aR&name=reproducibility}{reproduction package} used for the experiments in this work to the community.

%% file: sections/02_background.tex
% NOTE: to add keyword highlighting, go to main.tex ~207 "morekeywords="
\begin{figure}[!ht]
\vspace*{-1mm}
  \centering % frame=tb for top/bottom line
\begin{lstlisting}[language=pdl,emph=actions,xleftmargin=.15\textwidth, xrightmargin=.15\textwidth]
text:
- role: tools
  text: ${ tools }
- "Out of 1400 participants, 400 passed the test. What percentage is that?\n"
- def: actions
  model: replicate/ibm-granite/granite-3.1-8b-instruct
  parser: json
  spec: [{ name: str, arguments: { expr: str }}]  
- if: ${ actions[0].name == "calc" }
  then:
    lang: python
    code: result = ${ actions[0].arguments.expr }
\end{lstlisting}
  \caption{\label{fig:example}Basic example of a PDL program.}
  \vspace*{-5mm}
\end{figure}

% AutoPDL is based on the Prompt Declaration Language (PDL)~\cite{vaziriPDLDeclarativePrompt2024}, a YAML-based
% prompt programming model.
This paper uses PDL~\cite{vaziriPDLDeclarativePrompt2024} as a representation for exploring the search space of programs.
PDL programs are declarative and combine human readability with ease of execution. They represent the composition of calls to LLMs and tools, abstracting away the plumbing necessary for such compositions.
The output of the optimizer is also a PDL program, rather than simple textual prompts, so it is fully executable and could be further refined by a developer. 

\autoref{fig:example} shows a simple PDL program that uses a tool to answer a query. 
PDL is based on the premise that interactions with an LLM are mainly for the purpose of generating data. So, it allows users to specify the shape of data to be generated in a declarative way (in YAML), and is agnostic of any programming language. The first line of \autoref{fig:example} specifies that we are creating some text. Next, the first block in the itemized list defines the tools prompt. Line 3 contains a use of variable {\em tools}, expressed as a Jinja expression. This variable is
defined as the JSON Schema specification of a calculator tool (not shown in this figure, for the full program see Appendix~\ref{tools-calling}). Line 4 is the user query prompt. We do not specify the role explicitly as \emph{user} is the default role for prompts. Lines 5 through 8 show a model call. In PDL, the background context is accumulated implicitly, so the output of all blocks executed so far will be passed to the LLM as a list of input messages. The result of the model call is assigned to the variable {\em actions} (line 5). The model to be called is specified on line~6 (PDL is based on LiteLLM,\footnote{\url{https://github.com/BerriAI/litellm}} so this is a LiteLLM model id).
% \href{https://github.com/BerriAI/litellm}{LiteLLM}
Finally, lines 7 and 8 say that we parse the output of the model as JSON and type-check it according to the type on line 8.
Furthermore, when the inferencing server supports it, model calls with a schema use constrained decoding~\cite{willard_louf_2023}, enforcing syntactically correct JSON.\footnote{PDL additionally makes use of the heuristic \texttt{json-repair} package.}

On line 9, an if-statement checks whether the output of the LLM asks for the calculator tool.
If so, we use a Python code block to compute the requested tool call (lines 11 and 12). When we execute this program using the PDL interpreter, we obtain all the model inputs, followed by the model output, and finally the output of the tool call.
PDL has a rich set of control structures to allow writing a variety of prompting patterns, as well as functions to support libraries. For instance, \autoref{fig:pdlReact} shows a function call on line 4. In this paper, we consider the problem of automatically tuning prompts and choosing  prompting patterns for a given dataset. The following section explains our approach in further detail.

%% file: sections/03_approach.tex
\begin{figure}[!ht]
\vspace*{-1mm}
\centerline{\includegraphics[width=0.95\textwidth]{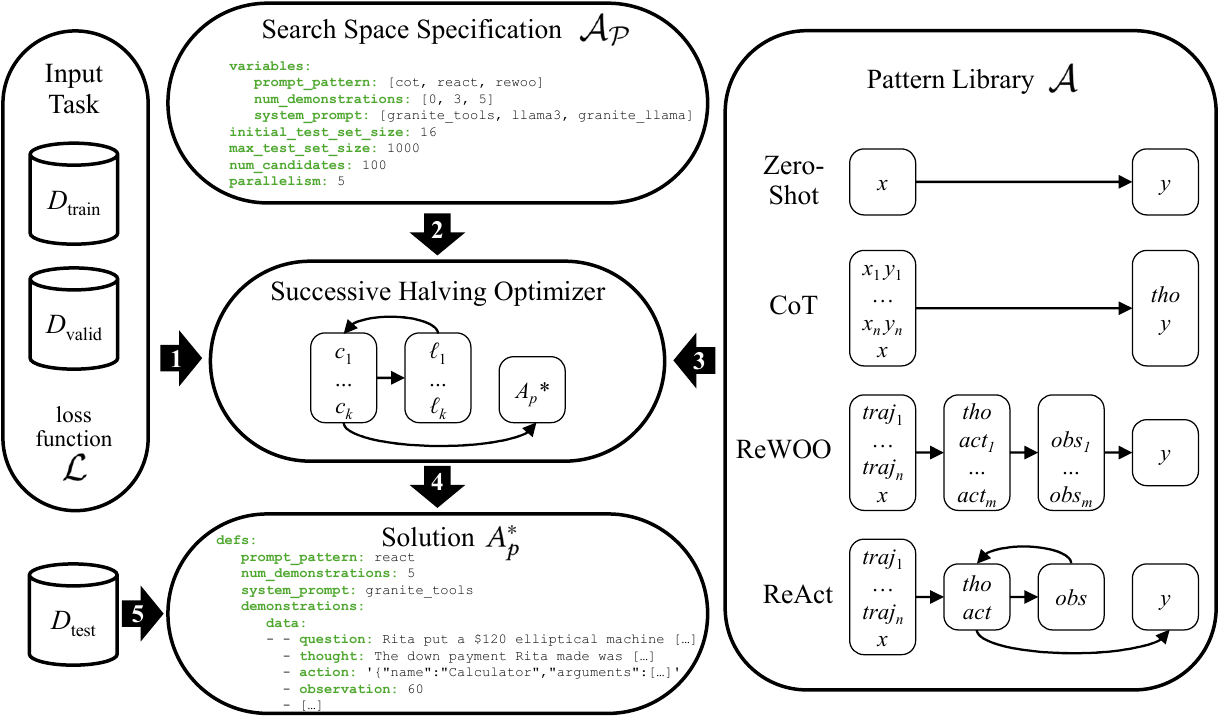}}	\caption{\label{fig:overview}Overview of our approach.}
    \vspace*{-5mm}
\end{figure}
\autoref{fig:overview} gives an overview of our approach.
Referring to the numbers in the arrows:

(1)~The input task is given by two disjoint datasets
$D_\mathrm{train}$ and $D_\mathrm{valid}$ and a loss function
$\mathcal{L}$.
The datasets comprise $\langle x,y\rangle$ instances, where $x$ is a
question, $y$ is the corresponding answer, and both are text strings.
The loss function evaluates the quality of an answer~$y$.
(2)~The search space specification $\mathcal{A}_\mathcal{P}$ is a YAML
file with the optimization variables and their possible values,
along with some hyperparameters.
For example, \lstinline[language=myyaml]{num_demonstrations: [0, 3, 5]}
indicates that each candidate will have zero, three, or five ICL
samples randomly drawn from~$D_\mathrm{train}$.
In the case of zero demonstrations, this is equivalent to the zero-shot baseline. If zero is an option, we bias our candidate sampling to always include one zero-shot candidate, just in case that baseline turns out to be the best-performing configuration.

(3)~The pattern library consists of four PDL functions.
Zero-shot is a baseline that simply prompts the LLM with $x$
and expects it to return~$y$.
CoT refers to
chain-of-thought~\cite{weiChainThoughtPromptingElicits2023} with
in-context learning~\cite{brown_et_al_2020}: the input includes a few
$x_iy_i$ pairs before the actual question~$x$, and the output includes
some reasoning thought $\mathit{tho}$ before the actual answer~$y$.
ReWOO~\cite{xuReWOODecouplingReasoning2023} refers to reasoning
without observations.
Here, the few-shot samples are trajectories $\mathit{traj}_i$.
A trajectory $\mathit{traj}_i$ consists of steps for a particular example problem instance in $D_\mathrm{train}$, \eg in the case of ReWOO, $\mathit{tho}$ and $\mathit{act}$ steps and their contents. In contrast to $x_iy_i$ pairs, $\mathit{traj}_i$ may contain many $\mathit{tho}$, $\mathit{act}$ and, depending on pattern, $\mathit{obs}$ steps, before reaching the solution~$y_i$.
The
first LLM call in ReWOO generates one reasoning thought $\mathit{tho}$ and multiple actions
$\mathit{act}_i$.
The PDL code executes each of the actions as a tool call to obtain the corresponding observations $\mathit{obs}_i$.
A final model call generates the answer $y$ based on the observations.
Finally, the ReAct pattern~\cite{yao_et_al_2023} starts with
few-shot trajectory examples $\mathit{traj}_i$ (brief example visible under Solution $A_p^*$ in \autoref{fig:overview}) and the question~$x$,
and then enters a TAO (thought, action, observation) loop.
In each loop iteration, the LLM generates $\mathit{tho}$ and
$\mathit{act}$, then the PDL code executes the action as a tool call,
and feeds the tool output back as an observation $\mathit{obs}$.
A special Finish action breaks out of the loop to return the answer~$y$.

\begin{wrapfigure}{r}{0.4\textwidth}
  \vspace{-6.5mm}
  \centering
\begin{lstlisting}[language=pdl,emph=ANSWER]
text:
  - include: ../../tools.pdl
  - include: ../../ReAct.pdl
  - call: ${ react }
    def: ANSWER
    args:
      task: "Question: ${ question }"
      model: ${ model }
      tools: ${ tools }
      trajectories: ${ demonstrations }
  - "\nThe answer is ${ ANSWER.answer }"
\end{lstlisting}
  \caption{Basic example of PDL program using the ReAct pattern.}
  \label{fig:pdlReact}
  \vspace{-5mm}
\end{wrapfigure}

Once inputs (1), (2), and (3) are in place, the successive
halving optimizer runs in a loop.
It starts with a small subset $D_v\subset D_\mathrm{valid}$ and many
candidates \mbox{$\mathcal{C}=\{c_1,\ldots,c_k\}\subseteq\mathcal{A}_\mathcal{P}$}
sampled from the search space.
Each iteration uses $D_v$ to evaluate the
corresponding losses $\ell_1,\ldots,\ell_k$.
Then each iteration keeps the $\frac{k}{2}$ candidates with the smallest
$\mathcal{L}$ while doubling the size of the validation subset~$D_v$.
See \sref{sec:succ_halving} for the algorithm.
(4)~After the last iteration, the best remaining candidate is
the solution~$A_p^*$.
This solution is a set of PDL definitions with concrete values for the optimization
variables, e.g., in \autoref{fig:overview}, \lstinline[language=myyaml]{num_demonstrations: 5} and
\lstinline[language=myyaml]{demonstrations: ...}
a list of ReAct trajectories.
(5)~This program can be used on the test set $D_\mathrm{test}$.
For instance, \autoref{fig:pdlReact} shows a call to the ReAct
function that passes \lstinline[language=pdl,mathescape=false]|${demonstrations}| from~$A_p^*$ as an argument.

%% file: sections/04_methodology.tex
This section describes the datasets used, the tools available to agents, and our experimental setup, including how we construct agent trajectories that demonstrate tool use for each dataset.

\subsection{Datasets}
\label{sec:methodologyDatasets}
We selected datasets that are widely used in the literature, span diverse tools and domains, and are representative of tool categories frequently studied in prior work (\eg calculator, search, code execution). In our experiments, each dataset has three disjoint splits:
$D_\mathrm{train}$ to sample few-shot samples from,
$D_\mathrm{valid}$ to evaluate candidates during optimization, and
$D_\mathrm{test}$ to evaluate the final chosen solution upon completion of optimization.

\paragraph{GSM8K}
The Grade School Math~\cite{cobbeTrainingVerifiersSolve2021} dataset consists of 8,792 grade school math problems. We sample 1,024 problems without replacement from the train set to use as our $D_\mathrm{valid}$ set, and 1,024 from the test set (consisting of 1,319 samples) to use as our $D_\mathrm{test}$. This leaves a $D_\mathrm{train}$ of 6,449 problems. Each problem consists of a word problem $x$ such as \textit{``What is fifteen more than a quarter of 48?''}, a sequence $\mathit{tho}$ of reasoning steps, and finally a plain numeric answer $y$ following a special delimiter.
For the CoT prompt pattern, we include these reasoning steps directly in the demonstrations. We use a regular expression to extract the numerical answer from the model solution, and define a correct solution as an exact match to the ground truth answer.

\paragraph{GSM-Hard}
\textcite{gao_et_al_2023} introduce a derivative of GSM8K with variables randomly changed to large numbers, with 1,319 samples. Unfortunately, GSM-Hard had 132 samples where the ground truth was incorrect, and hence we excluded those samples. We split the single set into equally sized $D_\mathrm{valid}$ and $D_\mathrm{test}$ ($n=594$), and use the GSM8K training set (6,449 samples) described above for $D_\mathrm{train}$ (\emph{cross transfer}). We use the same correctness criterion as for GSM8K.

\paragraph{FEVER}
The Fact Extraction and VERification dataset~\cite{thorneFEVERLargescaleDataset2018} is a question-answering dataset structured around fact-checking. The original dataset contained 185,445 claims that are true, false, or unverifiable, and associated with human annotated supporting, refuting, or neutral sentences and their Wikipedia article of origin. We follow the widely used derivative of this benchmark in BIG-bench~\cite{srivastavaImitationGameQuantifying2023}, which reformulates it into a true-or-false task by removing unverifiable claims. We sample 1,024 claims from the train set as $D_\mathrm{valid}$ and 1,024 from the test set as $D_\mathrm{test}$. This leaves a $D_\mathrm{train}$ of 5,696 claims. BIG-bench also does not include the supporting or refuting sentences, which we recover by joining on the original dataset, for use \eg as CoT demonstrations. To assess correctness, we check for the presence of ``true'' and ``false'' in the final line of the model response. If neither or both are present, we deem the response as incorrect, and otherwise the correctness is a direct match to the ground truth ``true'' or ``false''.

\paragraph{MBPP+}
MBPP~\cite{austinProgramSynthesisLarge2021} is a dataset of 974 mostly basic Python problems, with each example consisting of a natural language problem specification $x$ for a self-contained Python function~$y$,
% \eg set intersection
along with a single test case. Each problem has an extended set of test cases used for evaluation, which are not shown to the model.
\textcite{liuYourCodeGenerated2023} found that
MBPP test cases are incomplete, allowing proposed solutions to pass as correct, despite not matching the problem specification. Therefore, our experiments are based on MBPP+, which contains a subset of the problems, but a more complete set of test cases for each problem. We use these test cases to assess the correctness of the proposed solutions.
We use the 374 examples from the original MBPP dataset as $D_\mathrm{train}$, and split MBPP+ into $D_\mathrm{test}$ (224 samples) and $D_\mathrm{valid}$ (39 samples) based on which split they were in \emph{MBPP}. The discrepancy in number of MBPP+ samples is due to the exclusion of samples found in the MBPP train set, to avoid data contamination.
Our ReAct implementation follows \textcite{wangMINTEvaluatingLLMs2024}. We do not implement ReWOO as it is not reactive, \ie cannot include execution feedback.

\subsection{Tools}
Our prompt library represents actions, \ie tool calls, following the JSON tool calling schema~\cite{abdelaziz_et_al_2024}. An action is represented as a JSON object with \texttt{name} and \texttt{arguments} mapping. For example,
\lstinline[language=myyaml]|{"action": "Calc", "arguments": {"expr": "48/4"}}|
represents a call to a calculator with the expression to evaluate.
The PDL functions implementing the patterns (see Figure~\ref{fig:overview}) accept a list of tool definitions as an argument. Each element of that list is itself a \texttt{PDL} function. 
As both the agents and tools are implemented in PDL, the set of tools for a given task could itself be made a search space dimension, which we leave to future work.

\paragraph{Calculator}
For math datasets, we give the agentic approaches access to the \emph{Calc} tool. This tool evaluates a cleaned (\eg replacement of \verb|^| with $**$) expression with \href{https://github.com/sympy/sympy}{SymPy}, returning the result. In case of error, the function returns a warning that the expression was invalid, which may help the agent recover from invalid input.

\paragraph{Search}
For fact verification, we provide access to the \emph{Search} tool, which returns the summary of the first Wikipedia search result for the query. If no results are found, a hint to try again is returned, or if the title is too ambiguous, a list of possible disambiguations.

\paragraph{Execute}
We implement a programming agent for the code generation task, which can execute arbitrary code surrounded in XML-style \texttt{<execute>} tags. This tool executes the code in a Python shell, which returns the result of the final expression. This allows the agent to test its proposed solution against the given test case before submitting its solution.

\paragraph{Finish}
The most basic action is the \emph{Finish} action, or \texttt{<solution>} tag for the coding agent. This ends the agent's trajectory and results in the agent returning the value as the solution.

\subsection{Experimental Setup}
\label{sec:expSetup}
We evaluate the efficacy of our approach by running an optimization process to completion for each model \& dataset pair, and subsequently comparing the task accuracy. As a baseline, we evaluate each model in a zero-shot setting. This setting reflects the minimal effort approach by a user or developer, where they do not include any demonstrations with their query or task. As no model we investigate was specifically trained to create agentic trajectories in a zero-shot setting, it is not feasible to create a zero-shot setting under ReAct or ReWOO. % prompting techniques.

For the optimization process, we used an initial candidate set $\mathcal{C}$ of 100 candidate prompt configurations $c_i=\langle A,p\rangle\in\mathcal{C}$
%(configurations of demonstrations \& variable values)
per experiment. We fixed the size of the initial validation subset $d_\mathrm{valid} \subseteq D_\mathrm{valid}$ to 16. %, and the reduction factor $\eta$ to~2.
We define $\mathcal{L}(c_i,\, d_\mathrm{valid}) = -\text{Accuracy}(c_i,\, d_\mathrm{valid})$ for a given candidate~$c_i$, where accuracy is the fraction in $d_\mathrm{valid}$ of correctly solved problems by $c_i$ as per the dataset definition of correctness. % When both zero demonstrations and the CoT prompt pattern are included in the search space, we explicitly bias the candidate sampler to include at least one configuration with CoT and zero demonstrations. This guarantees the presence of a zero-shot baseline in the candidate pool, ensuring that the resulting optimized program performs at least as well as the baseline. 
% Additionally, a variable \textbf{Demonstrations} that holds the demonstrations or trajectories, defined as $D_t \subset D_{\text{train}}$ where $|D_t| = $ number of demonstrations, is sampled with replacement for each candidate.
For each candidate $c_i$, $p=\langle n,d_\textrm{train},\textrm{instr}\rangle\in\mathcal{P}$ where $n$ few-shot samples or trajectories $d_\mathrm{train}\in (D_\mathrm{train})^n$ are randomly sampled with replacement.
% TODO: can we reuse the $n$ and $d_\mathrm{train}$ notation introduced in the introduction for this? $p=\langle n,d_\textrm{train},\textrm{instr}\rangle\in\mathcal{P}$
The number of possible values for $\mathcal{P}$ is combinatorially large and depends on the dataset $D_\mathrm{train}$ used. Finally, upon completion of an optimization process, the optimal candidate is evaluated on~$D_{test}$.
%These values are reported in~\autoref{tab:mainResultsOpt}.
%, potentially yielding thousands of distinct subsets. 
% In this configuration, the combinatorial space follows $N_{\text{total}} = (\#\text{Prompt patterns}) \times (\#\text{Num. demonstrations}) \times (\#\text{System prompts}) \times \sum_{k \in \{\text{Num. demonstrations}\}} |D_{\mathrm{train}}|^k$. For example, a $|D_{\mathrm{train}}| =$ 6,000 would result in a search space of $\approx 2.100 \times 10^{20}$ configurations. 

\paragraph{Constructing Trajectories}
As we are also optimizing over agentic trajectories, we also need a set of trajectories to sample few-shot samples from. To achieve this, we create a basic agentic trajectory $\mathit{traj}_i$ for each training example $\langle x_i,y_i\rangle$, following a rule-based transformation. We design and apply a template to each dataset, which is relatively simple and easy to implement for other datasets (details are provided in \sref{sec:trajectory_construction}). Prior work has introduced approaches to bootstrapping trajectories \eg in software engineering~\cite{panTrainingSoftwareEngineering2024a}, tool use demonstrations~\cite{start}, and reasoning paths~\cite{zelikmanSTaRBootstrappingReasoning2022}, which could be applied to this problem. While we acknowledge the shortcomings of manual template construction, we argue this approach has two strengths: it is generalizable in the sense that templates can be mixed and matched, and that the trajectories are directly based on the datasets used.
%without \eg noise from a bootstrapping LLM.
Additionally, we wanted to work with commonly used datasets that cover a variety of tools and domains, rather than emerging datasets containing trajectories or tool use demonstrations.

\paragraph{Models}
We aim to study models of various abilities, \eg natural language or code, various creators, and various sizes, ranging from single digit billions of parameters up to the edge of feasibility on consumer hardware. We include seven models available on inference service IBM watsonx\footnote{\url{https://www.ibm.com/watsonx}} in our study. We select three generalist natural language instruction models, \textsc{Llama 3.1 8B}, \textsc{3.2 3B}, and \textsc{3.3 70B} from the open-source and widely studied LLaMa family~\cite{dubeyLlama3Herd2024}. We further select three models from~\textcite{mishraGraniteCodeModels2024}, which predate \textcite{dubeyLlama3Herd2024} by approximately 3 months. We select \textsc{Granite 13B Instruct V2} as a generalist model, and \textsc{Granite 20B} and \textsc{34B Code Instruct} as code models. We select \textsc{Granite 3.1 8B} as an additional generalist model~\cite{granite2024granite}. All of our experiments use greedy decoding, \ie no sampling, to limit the impact of hyperparameter choice. The number of models we evaluate is limited by cost in \$/token terms and execution time. By studying various models, we demonstrate the generalizability of our approach. 

\paragraph{Alternative Setups}
We evaluate two alternative experimental setups. First, to investigate low-resource scenarios, we examine whether performance on one dataset can be improved by using demonstrations $D_\mathrm{train}$ from a similar dataset, while optimizing w.r.t.\ $D_{\mathrm{valid}}$. For this experiment, we investigate whether performance on GSM-Hard can be improved by using demonstrations from \emph{GSM8K}, while optimizing w.r.t.\ GSM-Hard $D_{\mathrm{valid}}$.
Second, to explore saving optimization costs, we assess whether optimized prompt programs of one model can transfer well to a frontier model. The intuition is that while that might not be the best program for the frontier model, it might at least improve somewhat over the baseline. To this end, we evaluate the optimized PDL programs of \textsc{LLaMa 3.1 70B} on OpenAI's \texttt{gpt-4o-mini-2024-07-18}, for each dataset.

%% file: sections/05_results.tex
This section describes the results of our empirical study to evaluate our AutoPDL approach and answer the following research questions:

\xyz{To what extent does AutoPDL improve accuracy, and how much does the best solution vary by task and model?}
RQ1 asks to what degree our AutoPDL approach can improve model performance over their zero-shot baseline across a variety of commonly used benchmarks. We also seek to identify trends, if any, in optimal configurations, \eg whether more few-shots is always better, or whether certain prompt patterns are particularly suited to certain problem domains.

\xyz{Can AutoPDL make up for a missing few-shot example bank for a given task by reusing the example bank from a similar task?}
RQ2 investigates whether optimizing on one dataset using demonstrations from another, related, dataset can result in higher performance than using no demonstrations (zero-shot). This RQ addresses a low-resource scenario in which a limited pool of demonstrations exists in one dataset, but a dataset from a similar domain has a large pool.

\xyz{Do solutions found by AutoPDL improve performance on frontier models, even when optimized for open-source models?}
It can be expensive to run optimization against commercial frontier-model APIs. RQ3 assesses whether optimized prompt programs are transferable to different (and likely stronger) models than those they were optimized with.

\autoref{tab:mainResultsOpt} reports the results of our optimization and evaluation procedure.
We performed three complete optimization runs for FEVER, GSM8K, and MBPP+, and report mean accuracy, standard deviation in percentage points (\ie absolute, \emph{not} relative, uncertainty), and the pattern of the highest scoring run.
Across models and datasets, we generally find some improvement over the zero-shot baseline with few-shot chain-of-thought, or agentic patterns ReAct or ReWOO. % and \autoref{fig:accuracyLollipops} 

\begin{table}[!ht]
\vspace*{-1mm}
\centering
\caption{Model accuracies across datasets for baseline (zero-shot) and optimized versions. %Multiple runs reported for GSM8K and MBPP+, in which case ``Optimized'' Accuracy is reported as arithmetic mean, and ``Delta'' is the \emph{maximum} difference.% Optimal pattern (with number of demonstrations, and system prompt instructions for ReAct) denoted in parenthesis. Total optimization runtime reported in HH:mm format.
}
\label{tab:mainResultsOpt}
\begin{adjustbox}{width=0.8\textwidth}
\begin{tblr}{
  width=\linewidth, % tell tabularray what width our table should be
  colspec={l l r l r l c l}, % set column alignments
  vline{4} = {3-Z}{dotted}, % add dotted vline in col 4 from row 3 onwards
  hline{3,10,17} = {0.5pt}, % add hlines between datasets
  % hline{3,9,15} = {0.5pt}, % add hlines between datasets
  hline{1,Z} = {1pt}, % add hlines at top and bottom
  row{even} = {gray!5}, % color even rows gray
  row{1} = {white, font=\large\bfseries}, % set headers
  row{2} = {white, font=\small}, % set subheaders
  cell{10}{1} = {white},
  rowspec={Q|[gray]}, % set hline to gray under Accuracy
  cell{1}{3}={c=3}{c}, % set Accuracy col to width 3 & center
  cell{3}{1}={r=7}{c}, % set dataset cells to span 8 rows
  % cell{3}{1}={r=6}{c}, % set dataset cells to span 6 rows
  cell{10}{1}={r=7}{c}, % set dataset cells to span 8 rows
  % cell{9}{1}={r=6}{c}, % set dataset cells to span 6 rows
  cell{17}{1}={r=7}{c}, % set dataset cells to span 8 rows
  % cell{15}{1}={r=6}{c}, % set dataset cells to span 6 rows
  cell{1}{1,2,6,7}={r=2}{c}, % set non-Accuracy headers to span 2 rows
  % note{a} = {COT: Cumulative Operating Time, in hours},
}
Dataset & Model & Accuracy & & & Best Pattern & Runtime \\
 &  & Zero-Shot & Optimized & Delta & & \\
FEVER & Granite 3.1 8B & \SI{72.9}{\percent} & \SI{76.4 \pm 3.3}{\percent} & +3.5pp & ReWOO (5 shot) & 13:53 \\
FEVER & Granite 13B Instruct V2 & \SI{6.5}{\percent} & \SI{74.0 \pm 1.4}{\percent} & +67.5pp & ReWOO (3 shot) & 08:28 \\
FEVER & Granite 20B Code & \SI{39.7}{\percent} & \SI{63.1 \pm 1.6}{\percent} & +23.4pp & CoT (3 shot) & 12:03 \\
FEVER & Granite 34B Code & \SI{56.4}{\percent} & \SI{62.6 \pm 3.8}{\percent} & +6.2pp & CoT (3 shot) & 10:07 \\
FEVER & LLaMA 3.1 8B & \SI{68.5}{\percent} & \SI{77.5 \pm 0.8}{\percent} & +9.0pp & CoT (3 shot) & 05:06 \\
FEVER & LLaMA 3.2 3B & \SI{38.0}{\percent} & \SI{66.3 \pm 0.9}{\percent} & +28.3pp & ReWOO (5 shot) & 10:10 \\
FEVER & LLaMA 3.3 70B & \SI{67.6}{\percent} & \SI{78.1 \pm 0.6}{\percent} & +10.5pp & ReWOO (3 shot) & 21:27 \\
GSM8K & Granite 3.1 8B & \SI{74.2}{\percent} & \SI{74.2 \pm 0.6}{\percent} & +0.0pp & Zero-Shot (Baseline) & 08:56 \\  % !!!
GSM8K & Granite 13B Instruct V2 & \SI{23.0}{\percent} & \SI{30.9 \pm 1.0}{\percent} & +7.9pp & CoT (3 shot) & 09:20 \\
GSM8K & Granite 20B Code & \SI{68.7}{\percent} & \SI{68.7 \pm 0.1}{\percent} & +0.0pp & Zero-Shot (Baseline) & 09:27 \\
GSM8K & Granite 34B Code & \SI{72.1}{\percent} & \SI{72.1 \pm 0.1}{\percent} & +0.0pp & Zero-Shot (Baseline) & 08:52 \\
GSM8K & LLaMA 3.1 8B & \SI{78.4}{\percent} & \SI{85.3 \pm 0.6}{\percent} & +6.9pp & CoT (5 shot) & 08:48 \\
GSM8K & LLaMA 3.2 3B & \SI{71.8}{\percent} & \SI{75.3 \pm 0.4}{\percent} & +3.5pp & CoT (3 shot) & 16:36 \\
GSM8K & LLaMA 3.3 70B & \SI{85.5}{\percent} & \SI{95.4 \pm 0.2}{\percent} & +9.9pp & CoT (3 shot) & 07:50 \\
MBPP+ & Granite 3.1 8B & \SI{62.9}{\percent} & \SI{62.9 \pm 0.0}{\percent} & +0.0pp & Zero-Shot (Baseline) & 02:14 \\
MBPP+ & Granite 13B Instruct V2 & \SI{10.7}{\percent} & \SI{19.2 \pm 1.2}{\percent} & +8.5pp & ReAct (5 shot) & 04:02 \\
MBPP+ & Granite 20B Code & \SI{51.8}{\percent} & \SI{51.8 \pm 0.4}{\percent} & +0.0pp & Zero-Shot (Baseline) & 03:43 \\
MBPP+ & Granite 34B Code & \SI{48.7}{\percent} & \SI{61.3 \pm 1.0}{\percent} & +12.6pp & ReAct (3 shot) & 04:54 \\
MBPP+ & LLaMA 3.1 8B & \SI{61.2}{\percent} & \SI{62.8 \pm 4.0}{\percent} & +1.6pp & ReAct (5 shot) & 01:45 \\
MBPP+ & LLaMA 3.2 3B & \SI{58.0}{\percent} & \SI{58.0 \pm 0.4}{\percent} & +0.0pp & Zero-Shot (Baseline) & 02:01 \\
MBPP+ & LLaMA 3.3 70B & \SI{71.4}{\percent} & \SI{71.4 \pm 0.0}{\percent} & +0.0pp & Zero-Shot (Baseline) & 02:27 \\
\end{tblr}
\end{adjustbox}
\vspace*{-1.5mm}
\end{table}

\paragraph{FEVER}{We observed the minimum improvement in \textsc{Granite 3.1 8B}, with a 3.5 percentage point~(pp) improvement, and a maximal improvement of 67.5pp for \textsc{Granite 13B Instruct V2}. In terms of prompt pattern, CoT and ReWOO are equally represented. ReAct was not the optimal for any of the models. Interestingly, the largest model (\textsc{LLaMa 3.3 70B}) benefited by 10.5pp from 3-shot ReWOO. FEVER runtimes are generally higher than the other benchmarks, likely due to the large number of tokens involved by including Wikipedia content.}

\paragraph{GSM8K}{The highest improvement recorded (9.9pp) was for \textsc{LLaMa 3.3 70B} using 3-shot CoT, while the minimum improvement of 3.5pp was in \textsc{LLaMa 3.2 3B} using 3-shot CoT. ReAct and ReWOO were not the optimal for any model. For \textsc{Granite 3.1 8B}, \textsc{Granite 20B Code}, and \textsc{Granite 34B Code}, no improvement over the zero-shot baseline was identified. This was somewhat surprising, as generally including even some few-shot samples improves performance in LLMs.} % cite

\paragraph{MBPP+}{Several models benefited from execution feedback, as 3 out of 7 had ReAct as the optimal prompt pattern (ReWOO was excluded as described in \sref{sec:expSetup}). The greatest improvement of 12.6pp was in \textsc{Granite 34B Code}, and 8.5pp in \textsc{Granite 13B Instruct V2}, likely due to its poor programming performance as a generalist, non-code model. In contrast, the smaller \textsc{LLaMa 3.1 8B} model had high zero-shot performance of 61.2\%, yet still improved by up to 6.2pp (1.6pp on average) with ReAct. No improvement was observed for \textsc{Granite 3.1 8B}, \textsc{Granite 20B Code}, or the other \textsc{LLaMa} models.}

% \begin{findingsbox}
% \textbf{RQ 1}: AutoPDL generally resulted in accuracy boosts of 6--30\% over the zero-shot baseline, with a maximum improvement of 68.9pp for \textsc{Granite 13B Instruct V2} on the FEVER benchmark. No single prompt pattern was the best within a model or dataset group.
% %, suggesting that our approach benefits from jointly searching over prompt patterns and prompt content.
% \end{findingsbox}

\begin{table}[!ht]
\vspace*{-1mm}
\centering
\caption{Model accuracies on GSM-Hard for cross optimization experiment.}
\label{tab:mainResultsGsmHard}
\begin{adjustbox}{width=0.8\textwidth}
\begin{tblr}{
  width=\linewidth, % tell tabularray what width our table should be
  colspec={l l r l r l c l}, % set column alignments
  vline{4} = {3-Z}{dotted}, % add dotted vline in col 4 from row 3 onwards
  hline{3} = {0.5pt}, % add hlines between datasets
  hline{1,Z} = {1pt}, % add hlines at top and bottom
  row{even} = {gray!5}, % color even rows gray
  row{1} = {white, font=\large\bfseries}, % set headers
  row{2} = {white, font=\small}, % set subheaders
  rowspec={Q|[gray]}, % set hline to gray under Accuracy
  cell{1}{3}={c=3}{c}, % set Accuracy col to width 3 & center
  % cell{3}{1}={r=6}{c}, % set dataset cells to span 6 rows
  cell{3}{1}={r=6}{c}, % set dataset cells to span 8 rows
  cell{1}{1,2,6,7}={r=2}{c}, % set non-Accuracy headers to span 2 rows
  % note{a} = {COT: Cumulative Operating Time, in hours}, \TblrNote{a}
}
Dataset & Model & Accuracy & & & Best Pattern & Runtime \\
 &  & Zero-Shot & Optimized & Delta & \\
GSM-Hard & Granite 3.1 8B & \SI{36.0}{\percent} & \SI{37.8 \pm 0.8}{\percent} & +1.8pp & ReAct (5 shot, Granite Tools) & 23:34 \\
GSM-Hard & Granite 13B Instruct V2 & \SI{4.4}{\percent} & \SI{6.2 \pm 0.7}{\percent} & +1.8pp & CoT (5 shot) & 10:26 \\
GSM-Hard & Granite 20B Code & \SI{28.8}{\percent} & \SI{27.2 \pm 4.5}{\percent} & +0.0pp & Zero-Shot (Baseline) & 11:14 \\
GSM-Hard & Granite 34B Code & \SI{27.9}{\percent} & \SI{31.0 \pm 0.9}{\percent} & +3.0pp & CoT (3 shot) & 10:20 \\
GSM-Hard & LLaMA 3.2 3B & \SI{26.3}{\percent} & \SI{26.8 \pm 0.6}{\percent} & +0.5pp & CoT (5 shot) & 17:08 \\
GSM-Hard & LLaMA 3.3 70B & \SI{47.3}{\percent} & \SI{53.8 \pm 0.4}{\percent} & +6.5pp & CoT (5 shot) & 11:03 \\
\end{tblr}
\end{adjustbox}
\vspace*{-3mm}
\end{table}
\begin{table}[!ht]
\centering
\caption{Model accuracy for GPT-4o-mini cross experiment results}
\label{tab:mainResultsGpt4o}
\begin{adjustbox}{width=0.7\textwidth}
\begin{tblr}{
  width=\linewidth, % tell tabularray what width our table should be
  colspec={l l r l@{}r l}, % set column alignments {l l r l@{}r l} {X[l] X[l] X[r] X[l] X[r] X[l]}
  vline{4} = {3-Z}{dotted}, % add dotted vline in col 4 from row 3 onwards
  hline{3} = {0.5pt}, % add hlines between datasets
  hline{1,Z} = {1pt}, % add hlines at top and bottom
  row{even} = {gray!5}, % color even rows gray
  row{1} = {white, font=\large\bfseries}, % set headers
  row{2} = {white, font=\small}, % set subheaders
  rowspec={Q|[gray]}, % set hline to gray under Accuracy
  cell{1}{3}={c=3}{c}, % set Accuracy col to width 3 & center
  cell{1}{1,2,6}={r=2}{c}, % set non-Accuracy headers to span 2 rows
  % note{a} = {COT: Cumulative Operating Time, in hours}, \TblrNote{a}
}
Dataset & Model & Accuracy & & & Best Pattern \\
 &  & Zero-Shot & Optimized & Delta & \\
FEVER & GPT-4o-mini & \SI{83.7}{\percent} & \SI{87.7}{\percent} & +4.0pp & CoT (3 shot) \\
GSM-Hard & GPT-4o-mini & \SI{45.6}{\percent} & \SI{54.9}{\percent} & +9.3pp & ReAct (5 shot, Granite LLaMa) \\
GSM8K & GPT-4o-mini & \SI{77.8}{\percent} & \SI{90.9}{\percent} & +13.1pp & CoT (5 shot) \\
MBPP+ & GPT-4o-mini & \SI{72.3}{\percent} & \SI{72.3}{\percent} & +0.0pp & Zero-Shot (Baseline) \\
\end{tblr}
\end{adjustbox}
\vspace*{-5mm}
\end{table}

\paragraph{Missing Few-Shot Example Bank}{We optimized the PDL program for GSM-Hard, using GSM8K demonstrations, three times and report results in \autoref{tab:mainResultsGsmHard}. We found that in most cases, GSM8K demonstrations were at least not harmful for models on GSM-Hard, with up to 6.5pp improvement for \textsc{LLaMa 3.3 70B} using 5-shot CoT.}
% However, minimal or no improvement was observed for the other models.
% As one of the largest evaluated models, \textsc{LLaMa 3.3 70B} likely generalizes best and makes effective use of the calculator tool. Its newer counterpart, \textsc{LLaMa 3.3 70B} also sees an improvement of 6.1pp with 5-shot CoT.}

% table used to be here

\paragraph{Commercial Frontier Model}{To assess whether performance gains in one model can be achieved in another, we evaluate the optimized PDL programs of \textsc{LLaMa 3.1 70B} on OpenAI's \texttt{gpt-4o-mini-2024-07-18} and report results in \autoref{tab:mainResultsGpt4o} (we did not use \textsc{LLaMa 3.3 70B} here because we did this experiment earlier and did not have the time and resources to repeat it for the final version of this paper). For all dataset/prompt pattern pairs that resulted in improvement for \textsc{LLaMa 3.1 70B}, we found a surprising improvement in \textsc{GPT4o-mini} of at least 4pp on FEVER using 3-shot CoT, 9.3pp on GSM-Hard (using GSM8K demonstrations) with 5-shot ReAct (Granite LLaMa instructions), and up to 13.1pp on GSM8K using 5-shot CoT. This suggests that optimizing for an open-source model can also benefit a closed-source model.}

% \begin{findingsbox}
% \textbf{RQ 3}: Optimized prompt programs may generalize well between models, with an increase of up to 13.1pp by evaluating prompt programs optimized for \textsc{LLaMa 3.1 70B} on \texttt{gpt-4o-mini-2024-07-18}.
% \end{findingsbox}

%% file: sections/06_related.tex
The closest related work is on \textbf{prompt optimization}.
APE starts with an LLM-generated set of candidate prompts, then
performs rejection sampling based on evaluation on a subset of
data~\cite{zhou_et_al_2023_iclr}. ZOPO incorporates a Neural Tangent Kernel-based derived Gaussian process into standard zeroth-order optimization for an efficient search of a locally-optimal instruction~\cite{huLocalizedZerothOrderPrompt2024}. 
Unlike our approach, neither APE nor ZOPO optimize few-shot samples.
Aviary can also jointly optimize over prompt pattern, instruction, and few-shot examples~\cite{narayananAviaryTrainingLanguage2024}. However, it would require the definition of a custom operator.
CEDAR uses a demonstration pool, from which it retrieves few-shot
examples at query time~\cite{nashid_sintaha_mesbah_2023}.
Unlike our approach, these few-shot samples are retrieved on a
per-inference basis, not optimized ahead-of-time.

EASE leverages embeddings to represent few-shot examples, and uses a neural bandit algorithm to find an ordered set that performs well for test queries from a given task~\cite{wuPromptOptimizationEASE2024}. An extension of their approach jointly optimizes demonstrations and instructions. However, the approach requires both an additional embedding model, and the training of a new model to predict validation scores from embeddings. Unlike our approach, EASE does not optimize over agentic patterns.

DSPy optimizes instructions and few-shot samples for a chain of LLM
calls~\cite{khattab_et_al_2024} (not just a single call like APE or
CEDAR).
Also, DSPy takes away control over the exact prompt from the
programmer, which our approach preserves.
Similarly to DSPy, TextGrad also optimizes a chain of LLM calls, by
using LLMs to back-propagate modifications to instructions in
prompts~\cite{yuksekgonul_et_al_2024}. However, unlike our approach, neither of these optimize agentic patterns.
%

%Black-box optimization approaches attempt to optimize user prompts to best suit the LLM, without updating model parameters.
BPO trains a sequence-to-sequence model on prompts augmented by an LLM incorporating human preferences, producing a model that improves given input prompts~\cite{chengBlackBoxPromptOptimization2024}. APOHF introduce a strategy to select a pair of prompts to query the user for preference feedback, which they use to optimize LLM-generated instructions on a validation set~\cite{linPromptOptimizationHuman2024}. However, neither of these approaches explicitly optimize demonstrations or agentic patterns.
% But it does not optimize agentic patterns, which our approach does.
%
% Two papers on optimizing agents appeared in July 2024.
%
EvoAgent optimizes the instructions of a population of agents via
crossover, mutation, and selection~\cite{yuan_et_al_2024}.
It then forms an ensemble from the final, fittest, population.
GPTSwarm represents each agent as a graph, then freezes intra-agent
edges and optimizes the placement of additional inter-agent
edges~\cite{zhuge_et_al_2024}.
Unlike our approach, neither Evo\-Agent nor GPTSwarm optimize the
agentic pattern inside individual agents, nor do they optimize
few-shot samples.

% AutoSklearn \cite{feurer_et_al_2015} dec
% - Bayesian
% DAUB \cite{sabharwal_samulowitz_tesauro_2016} feb
% - incremental data allocation
% successive halving \cite{jamieson_talwalkar_2016} may
% - fast randomized, incremental data allocation
% TPOT \cite{olson_et_al_2016} mar
% - genetic
% AlphaD3M \cite{drori_et_al_2018} jul
% - MCTS, RL

Another closely related field of study is \textbf{AutoML}. 
%, or
% automated machine learning.
%
Auto-sklearn~\cite{feurer_et_al_2015} used Bayesian optimization to
jointly perform both algorithm selection and hyperparameters of a
scikit-learn pipeline~\cite{buitinck_et_al_2013}.
While different, we see some analogy between algorithms and agentic
patterns, and between hyperparameters and few-shot samples.
DAUB first evaluates many candidate models on a small amount of data,
then successively reduces candidates and increases data to ultimately
pick a strong model~\cite{sabharwal_samulowitz_tesauro_2016}.
The successive-halving algorithm takes a similar
approach~\cite{jamieson_talwalkar_2016}.
Our approach is inspired by the same incremental data allocation idea.
While both randomized search and Bayesian optimization are popular in
AutoML, there are also more intricate approaches.
For instance, TPOT uses genetic algorithms~\cite{olson_et_al_2016},
and AlphaD3M uses Monte-Carlo tree search~\cite{drori_et_al_2018}.
We chose to start with a simpler technique that depends less on a
well-behaved optimization space.
That said, exploring more advanced AutoML optimizers could be fruitful future work for AutoPDL.
Lale~\cite{baudart_et_al_2021} treats AutoML as a source-to-source optimization, similar to this paper, but unlike AutoPDL, it has not been used to optimize agentic patterns or prompts.

%% file: sections/07_conclusion.tex
We present our AutoPDL approach for jointly optimizing prompting patterns and textual prompts for large language models, addressing the challenges associated with manual prompt engineering. By formulating the optimization as a discrete search over both agentic and non-agentic patterns, combined with instructions and few-shot samples, we leveraged successive halving to efficiently navigate this search space. Our evaluation across various datasets (FEVER, GSM8K, GSM-Hard, and MBPP+) and multiple models (from the LLaMA, Granite, and GPT families) demonstrates substantial accuracy improvements, up to 67.5 percentage points, and affirms that no single prompting strategy universally outperforms others across tasks and models. Additionally, generating code in a YAML-based prompt programming language (PDL) makes it executable, easy to modify, and readable by humans, supporting practical adoption and adaptation.

%% file: sections/08_appendix.tex
% \appendix
\section{Supplemental Material}
\subsection{Concrete Prompt Example}
\crefname{appendixfigure}{Appendix figure}{Appendix figures}
\crefalias{figure}{appendixfigure}
\setcounter{figure}{0}
\renewcommand{\thefigure}{A\arabic{figure}}

\setcounter{table}{0}
\renewcommand{\thetable}{A\arabic{table}}
% \crefname{figure}{Appendix Figure}{Appendix Figures}
% \Crefname{figure}{Appendix Figure}{Appendix Figures}
% \crefname{appendixfigure}{Appendix Figure}{Appendix Figures}
% \Crefname{appendixfigure}{Appendix Figure}{Appendix Figures}

\begin{figure}[!h]
  % \centering
  \begin{adjustbox}{width=0.9\textwidth,center}
\begin{lstlisting}[language=pdl,emph=ANSWER,xleftmargin=.2\textwidth, xrightmargin=.2\textwidth]
defs:
  prompt_pattern: cot
  num_demonstrations: 2
  demonstrations:
    data:
    - question: Tricia is a third of Amilia's age and Amilia is a quarter of Yorick's age. Yorick is twice 
        Eugene's age and Khloe is a third of Eugene's age. Rupert is 10 years older than Khloe but 2 years 
        younger than Vincent who is 22 years old. How old, in years, is Tricia?
      reasoning: |-
        Rupert is younger than Vincent by 2 years, so he is 22 years old - 2 years = <<22-2=20>>20 years old.
        Khloe is 10 years younger than Rupert so she is 20 years old - 10 years = 10 years old.
        Eugene is 3 times older than Khloe so he is 10 years old * 3 = <<10*3=30>>30 years old.
        Yorick is twice Eugene's age so he is 30 years old * 2 = <<30*2=60>>60 years old.
        Amilia is a quarter of Yorick's age so she is 60 years old / 4 = <<60/4=15>>15 years old.
        Tricia is a third of Amilia's age so she is 15 years old / 3 = <<15/3=5>>5 years old.
      answer: '5'
    - question: Emmalyn decided to paint fences in her neighborhood for twenty cents per meter. If there were 
        50 fences in the neighborhood that she had to paint and each fence was 500 meters long, calculate  
        the total amount she earned from painting the fences.
      reasoning: |-
        The total length for the fifty fences is 50*500 = <<50*500=25000>>25000 meters.
        If Emmalyn charged twenty cents to paint a meter of a fence, the total income she got from painting 
        the fences is $0.20*25000 =$5000
      answer: '5000'
  instruction: Answer the questions to the best of your abilities.
text:
  - include: CoT.pdl
  - "${ instruction }\n\n"
  - call: ${ chain_of_thought }
    args:
      examples: ${ demonstrations }
      question: ${ question }
      model: ${ model }
\end{lstlisting}
  \end{adjustbox}
  \caption{Basic example of a concrete prompt configuration in PDL, using CoT pattern and two demonstrations.}
  \label{fig:pdlConcrete}
\end{figure}

\begin{figure}[!h]
  % \centering
  \begin{adjustbox}{width=0.9\textwidth,center}
\begin{lstlisting}[language=pdl,emph=ANSWER,xleftmargin=.2\textwidth, xrightmargin=.2\textwidth]
Answer the questions to the best of your abilities.

Question: Tricia is a third of Amilia's age and Amilia is a quarter of Yorick's age. Yorick is twice Eugene's 
age and Khloe is a third of Eugene's age. Rupert is 10 years older than Khloe but 2 years 
younger than Vincent 
who is 22 years old. How old, in years, is Tricia?
Answer: Let's think step by step. Rupert is younger than Vincent by 2 years, so he is 22 years 
old - 2 years = <<22-2=20>>20 years old.
Khloe is 10 years younger than Rupert so she is 20 years old - 10 years = 10 years old.
Eugene is 3 times older than Khloe so he is 10 years old * 3 = <<10*3=30>>30 years old.
Yorick is twice Eugene's age so he is 30 years old * 2 = <<30*2=60>>60 years old.
Amilia is a quarter of Yorick's age so she is 60 years old / 4 = <<60/4=15>>15 years old.
Tricia is a third of Amilia's age so she is 15 years old / 3 = <<15/3=5>>5 years old.
The answer is 5

Question: Emmalyn decided to paint fences in her neighborhood for twenty cents per meter. If there were 50 
fences in the neighborhood that she had to paint and each fence was 500 meters long, calculate the total 
amount she earned from painting the fences.
Answer: Let's think step by step. The total length for the fifty fences is 50*500 = <<50*500=25000>>25000 meters.
If Emmalyn charged twenty cents to paint a meter of a fence, the total income she got from painting the 
fences is $0.20*25000 =$5000
The answer is 5000

Question: ${ question }
Answer: Let's think step by step.
\end{lstlisting}  
  \end{adjustbox}
\caption{Corresponding rendered prompt configuration for \autoref{fig:pdlConcrete}. With the exception of the non-rendered \lstinline[language=pdl]{\$\{ question \}} variable, this is the input to the language model.}
  \label{fig:pdlConcreteTxt}
\end{figure}
See~\autoref{fig:pdlConcrete} and~\autoref{fig:pdlConcreteTxt}.

\subsection{Tool Calling Code}
\label{tools-calling}

\begin{figure}[!h]
  \centering % frame=tb for top/bottom line
  
\begin{lstlisting}[language=pdl,xleftmargin=.2\textwidth, xrightmargin=.2\textwidth]
description: tool use
defs:
  tools:
    data:
    - name: calc
      description: Calculator function
      arguments:
        expr: 
          type: string
          description: Arithmetic expression to calculate
text:
- role: system
  text: You are Granite, developed by IBM. You are a helpful AI assistant 
    with access to the following tools. When a tool is required to answer 
    the user's query, respond with <|tool_call|> followed by a JSON list of 
    tools used. If a tool does not exist in the provided list of tools, 
    notify the user that you do not have the ability to fulfill the request.
  contribute: [context]
- role: tools
  text: ${ tools }
  contribute: [context]
- "Out of 1400 participants, 400 passed the test. What percentage is that?\n"
- def: actions
  model: replicate/ibm-granite/granite-3.1-8b-instruct
  parser: json
  spec: [{ name: str, arguments: { expr: str }}]  
- "\n"
- if: ${ actions[0].name == "calc" }
  then:
    lang: python
    code: result = ${ actions[0].arguments.expr }
\end{lstlisting}
  \caption{\label{fig:full}Basic example of a PDL program.}
\end{figure}

See ~\autoref{fig:full}.

% \newpage

\subsection{Optimization}\label{sec:succ_halving}

\begin{figure}[!h]
  % \begin{minipage}{.8\textwidth}
    \centering
    
    \begin{algorithm}[H]
      \caption{Successive Halving for PDL Optimization}
      \label{alg:succ_halving_loss_nor}
      \begin{algorithmic}[1]
        \REQUIRE Program candidate set \(\mathcal{C}\), validation dataset \(D_\mathrm{valid}\), initial validation subset size \(v_{\min}\), maximum validation subset size \(v_{\max}\)
        \STATE \(v \gets v_{\min}\)
        \WHILE{\(|\mathcal{C}| > 1\)}% \textbf{and} \(v \le |D_\mathrm{valid}|\)}
          \STATE \(d_\mathrm{valid} \gets \text{first } v \text{ elements of } D_\mathrm{valid} \text{ s.t. } d_\mathrm{valid} \subseteq D_\mathrm{valid}\) and \(|d_\mathrm{valid}| = v \)
          \FOR{each candidate \(c_i \in \mathcal{C}\)}
            \STATE Compute loss \(\ell_i \gets\mathcal{L}(c_i,\, d_\mathrm{valid})\)
          \ENDFOR
          \STATE \(\mathcal{C} \gets\) top \(\lfloor |\mathcal{C}|/2 \rfloor\) candidates with lowest loss
          \STATE \(v \gets \min(v_{\max}, 2 \cdot v)\)
        \ENDWHILE
        \RETURN Candidate in \(\mathcal{C}\) with lowest loss
      \end{algorithmic}
    \end{algorithm}
  % \end{minipage}
  \caption{Illustration of the Successive Halving algorithm used to optimize the PDL program by pruning poor candidates on progressively larger validation subsets.}
    \label{fig:succ_halving}
\end{figure}
\autoref{fig:succ_halving} describes our optimization algorithm, based on successive halving~\cite{jamieson_talwalkar_2016}. The algorithm accepts a candidate set sampled from possible configurations and demonstrations, a validation dataset to optimize against, an initial validation subset size, and a maximum validation subset size. AutoPDL allows the user to specify these options in a YAML configuration file, and ultimately saves its result as a PDL program. This source-to-source transformation enables the user to modify both the search space and the resulting optimized PDL program, allowing further modification and execution.

\subsection{Search Space}
The search space
% consists of a subset all configurations formed by
is the Cartesian product of the following discrete variables, each taking one value per candidate: % , itemjoin={{, }} itemjoin*={{, and }}
\begin{enumerate}[label=(\arabic*)]
\item $A\in\mathcal{A} = \{\textrm{Zero-Shot, CoT, ReWOO, ReAct}\}$, \ie the overall prompting pattern to apply.
\item \textbf{Number of demonstrations} $n\in \{0,\ 3,\ 5\}$. We selected these options as a representative sweep across no supervision, moderate few-shot use, and an upper-end case (in terms of token window). We limited the search space to three options to avoid combinatorial explosion and limit experiment cost. % TODO: refine
\item If $A = \textrm{ReAct}$, \textbf{System prompt} $\in \{$\texttt{Granite Tools}, \texttt{LLaMa 3}, \texttt{Granite LLaMa}$\}$. As the system prompt instructs the model how to format tool calls, it only has an effect on benchmarks with JSON tool calling (FEVER, GSM8K, and GSM-Hard) for candidates with the ReAct prompt pattern. We note that only for MBPP+, ReWOO is not included as a prompt pattern, and that we always include two trajectories displaying iterative refinement, \ie an example of a solution failing the example test case, followed by a passing solution, in line with \textcite{wangMINTEvaluatingLLMs2024}. This effectively increases the number of trajectories to $|traj|+2$.
\end{enumerate}

\subsection{Agent Trajectory Construction}\label{sec:trajectory_construction}
To optimize over agentic patterns and trajectories, we require a set of example trajectories to use during optimization. For this purpose, we create a basic agentic trajectory $\mathit{traj}_i$ for each training example $\langle x_i,y_i\rangle$, following a rule-based transformation outlined below.

\paragraph{GSM8K} 
To demonstrate tool use in ReAct, we instead derive a trajectory $\mathit{traj}$ as follows. We exploit the fact that there is at most one expression per reasoning step, by iterating through the steps. At each step, we append a `thought' to the trajectory, consisting of the text leading up to the math expression, concatenated with a reflection `I need to calculate'. We append a calculator tool call with the expression, and an `observation', \ie the result of the expression. Finally, we append a thought `The answer is ...', containing the ground truth answer, followed by the \emph{finish} action with the answer.
We follow the same procedure to create ReWOO trajectories, except we use slightly different wording, \eg `Calculate xyz' in place of `I need to calculate xyz', and omit the final thought and action. Additionally, we use string substitution to replace any assumed expression results in the trajectory with the corresponding variable.

\paragraph{FEVER}
To produce agent trajectories, we iterate over each article associated with a claim, append a thought `I need to search for ...', followed by the action, an observation containing the article summary, and finally a thought containing all the relevant sentences associated with that article for that claim, which we repeat for each article associated with a claim. This procedure is not ideal as there is no inherent order to the articles or sentences, even though there may be a natural ordering following the annotator's Wikipedia navigation. Finally, we append a thought `The claim is true/false' and the finish action, both with the ground truth answer. For chain-of-thought, we perform the same procedure except we only include the concatenated evidence sentences, as there is no tool use.

\paragraph{MBPP+}
To generate sample agent trajectories from the training set, we follow the agent pattern (without feedback) in-context examples by \textcite{wangMINTEvaluatingLLMs2024}, which consists of the problem~$x$, a thought such as ``The intersection is elements that are in both lists'', an execute action that contains proposed code \emph{and} an assertion calling the proposed method with the test case input from the prompt and comparing its output. This is then followed by an observation containing the execution result, \ie either \texttt{``[Executed Successfully with No Output]''} or a stack traceback. This allows the agent to iterate on solutions (up to five times in our implementation). %, until it submits its solution using the \texttt{<solution>} tag.
We use the full MBPP train set of 374 problems as $D_\mathrm{train}$, and split the MBPP+ dataset into $D_\mathrm{valid}$ and $D_\mathrm{test}$ based on problem id membership in MBPP, leaving 39 and 224 validation and test problems respectively.

To generate synthetic trajectories from the training set, we start with the natural language specification and single test case (the prompt), append the thought ``I should run a solution on the test case before proposing a solution.'', followed by the ground truth solution and substitute in the prompt test case following the pattern \texttt{[solution]res = ...; assert res == ..., ''Expected ... but got {}''.format(res)}. Subsequently, we append the observation \texttt{``[Executed Successfully with No Output]''}, the thought \texttt{``There is no more AssertionError. I can now submit the solution.''}, and finally the solution action with the ground truth solution. This naive approach allows us to provide demonstration trajectories, albeit simplistic ones that assume the first solution is correct. Sampling a reflection or thought from a strong model may be beneficial~\cite{start}, but we restrict our trajectories to rule based transformations. As ReWOO is not reactive, \ie without execution feedback, it does not make sense for MBPP. Hence, we exclude it from our experiments.

% \clearpage
\subsection{Results Plot}
\begin{figure}[!h]
  \centering
  \includegraphics[width=\linewidth]{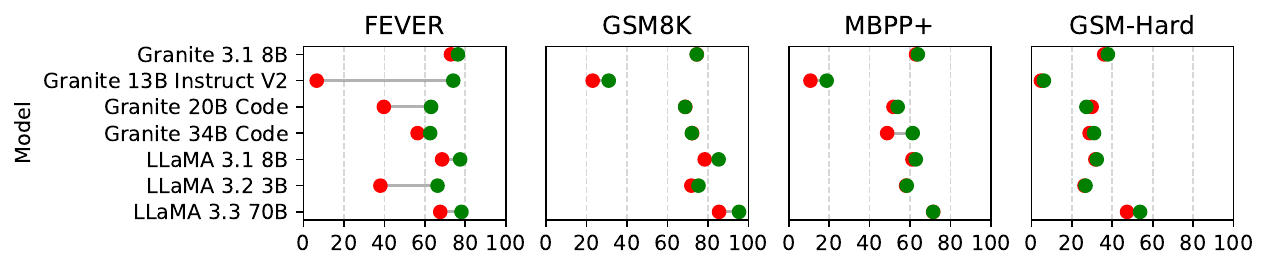}
  \caption{Comparison of optimized prompt program performance across models and datasets. Each barbell shows the accuracy improvement, if any, over the zero-shot baseline.}
  \label{fig:accuracyLollipops}
\end{figure}
In~\autoref{fig:accuracyLollipops}, we visualize the results from~\autoref{tab:mainResultsOpt} and~\autoref{tab:mainResultsGsmHard}.

\clearpage
\subsection{Accuracy \vs iterations}
\begin{figure}[!ht]
  \centerline{\includegraphics[width=0.8\textwidth]{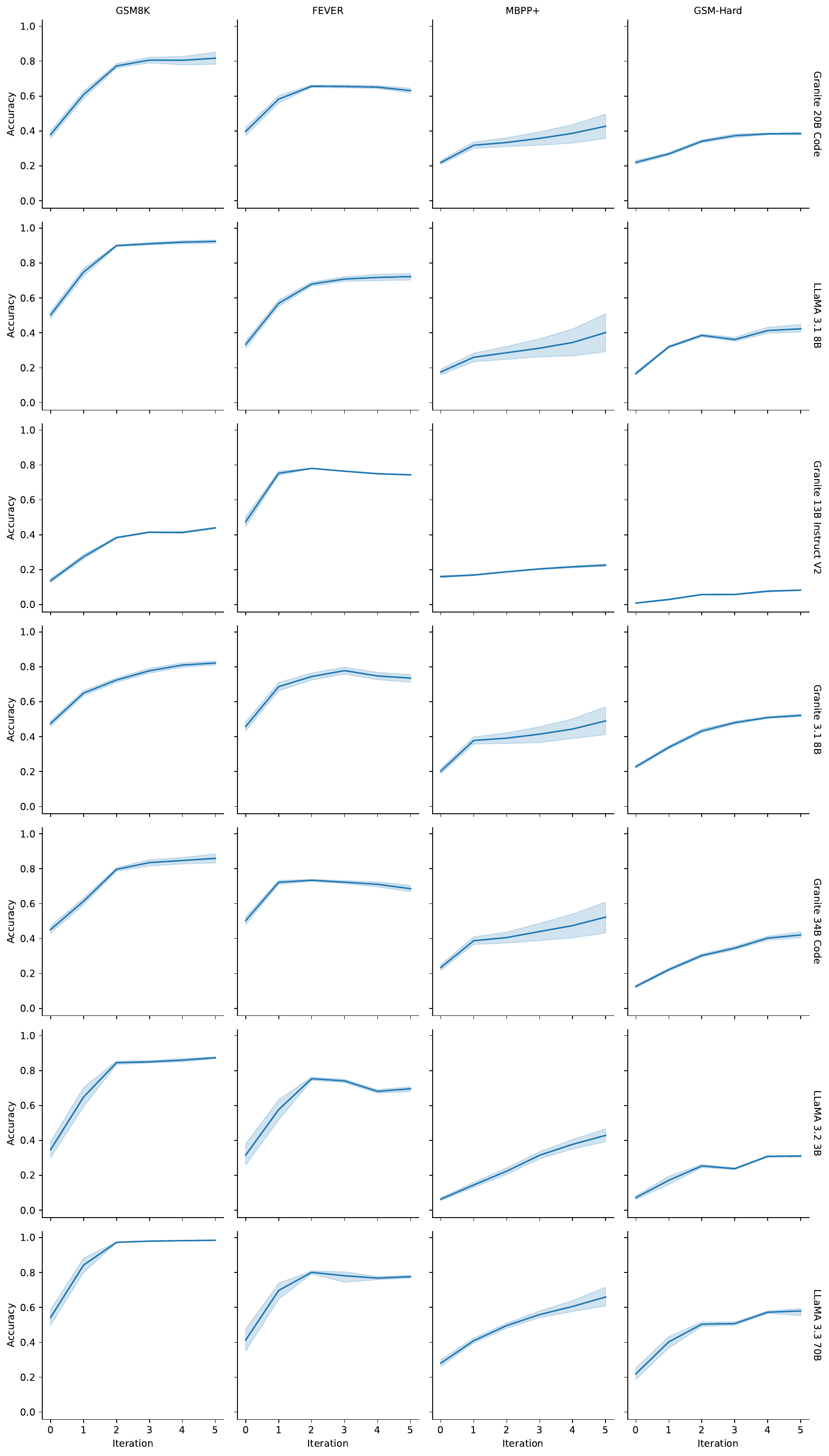}}
  \caption{\label{fig:iterations}Accuracy \vs iterations with $95\%$ confidence interval.}
\end{figure}
In~\autoref{fig:iterations}, we visualize the accuracy across candidates versus the iterations of the optimization process, including a $95\%$ confidence interval depicting the spread in accuracy across candidates. The confidence interval is computed using mean and 1,000 bootstraps. As the iterations increase, the number of candidates decreases, while the size of the validation set $D_v$ increases.